\documentclass[pdflatex,sn-mathphys-num]{sn-jnl}

\usepackage{graphicx}%
\usepackage{multirow}%
\usepackage{amsmath,amssymb,amsfonts}%
\usepackage{amsthm}%
\usepackage{mathrsfs}%
\usepackage{xcolor}%
\usepackage{textcomp}%
\usepackage{manyfoot}%
\usepackage{booktabs}%
\usepackage{algorithm}%
\usepackage{algorithmicx}%
\usepackage{algpseudocode}%
\usepackage{listings}%

\usepackage{CJKutf8}

\begin{document}

\title[From ``No'' to ``Know'': Challenges and Opportunities for Negation Understanding in Multimodal AI]{From No to Know: Taxonomy, Challenges, and Opportunities for Negation Understanding in Multimodal Foundation Models}


\author*[1]{\fnm{Mayank} \sur{Vatsa}}\email{mvatsa@iitj.ac.in}

\author[2]{\fnm{Aparna} \sur{Bharati}}\email{apb220@lehigh.edu}

\author[1]{\fnm{Surbhi} \sur{Mittal}}\email{mittal.5@iitj.ac.in}

\author[1]{\fnm{Richa} \sur{Singh}}\email{richa@iitj.ac.in}

\affil*[1]{\orgdiv{Srijan: COE on Generative AI}, \orgname{IIT Jodhpur}, \orgaddress{\country{India}}}

\affil[2]{\orgname{Lehigh University}, \orgaddress{\country{USA}}}


\abstract{Negation, a linguistic construct conveying absence, denial, or contradiction, poses significant challenges for multilingual multimodal foundation models. These models excel in tasks like machine translation, text-guided generation, image captioning, audio interactions, and video processing but often struggle to accurately interpret negation across diverse languages and cultural contexts. In this perspective paper, we propose a comprehensive taxonomy of negation constructs, illustrating how structural, semantic, and cultural factors influence multimodal foundation models. We present open research questions and highlight key challenges, emphasizing the importance of addressing these issues to achieve robust negation handling. Finally, we advocate for specialized benchmarks, language-specific tokenization, fine-grained attention mechanisms, and advanced multimodal architectures. These strategies can foster more adaptable and semantically precise multimodal foundation models, better equipped to navigate and accurately interpret the complexities of negation in multilingual, multimodal environments.}


\maketitle

\section{Introduction}\label{sec1}

\begin{figure}[h]
\centering
\includegraphics[width=\textwidth]{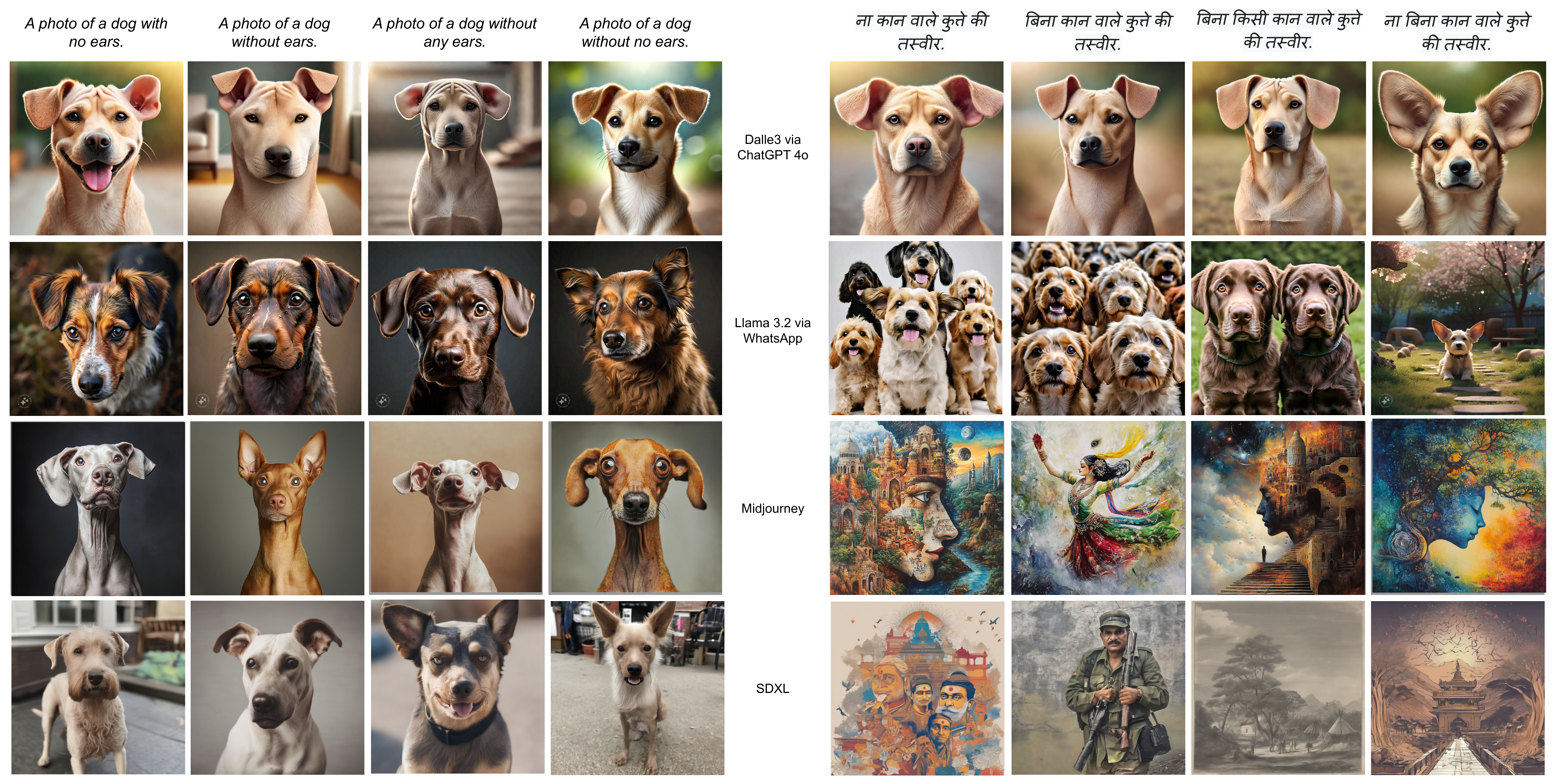}
           \caption{Text-to-image generative models face significant challenges in accurately interpreting negations within multilingual prompts. Regardless of the specific prompts, the models consistently produced images of dogs with ears, demonstrating a persistent inability to correctly process negated terms. This example illustrates the limitations of models such as DALL-E 3 and Llama 3.2 in handling various forms of negation across both English and Hindi languages. Furthermore, models like Midjourney and SDXL exhibit even more pronounced deficiencies, as they fail to process or understand the `Hindi’ language altogether.}
           \label{fig:fig1}
\end{figure}

Multimodal foundation models, including Llama, Gemini, and GPT-o1, rely on learning concepts within each modality and associating them across multiple modalities to perform tasks such as text-guided generation, editing, retrieval, image captioning, and audio/speech interactions. A common paradigm is to map image or visual features to text or language representations, either by learning a shared feature space between the two modalities~\cite{CLIP, BLIP} or by aligning one modality with the other~\cite{zhai2022lit, zhang2021vinvl}. While pre-training with vast amounts of paired data has led to emergent capabilities, complex prompts that require compositional understanding still remain a challenge for state-of-the-art multimodal foundation models.

Among the more complex linguistic constructs, \textit{negations} constitute a crucial phenomenon that varies widely across languages. Negations convey the absence, denial, or contradiction of a statement or action~\cite{khemlani2014negations}. Existing research indicates that negations within prompts frequently evade foundational models during various semantic tasks~\cite{truong-etal-2023-language, hsieh2023sugarcrepe, singh2024learn, patel2024tripletclip}. Language models, even those with robust syntactic and semantic capabilities, struggle with negations in neural translation~\cite{hossain2020s} and fill-in-the-blank tasks~\cite{kassner2020negated}. However, comprehending negations, though challenging for learning-based models~\cite{testoni2022artificial}, is essential for commonsense reasoning~\cite{safavi2021negater, schon2021negation} and highly desirable in image-text retrieval~\cite{patel2024tripletclip} and text-to-image generation settings~\cite{stablediff}, both of which rely on multimodal foundation models. Inaccurate handling of negation in these contexts can lead to significant real-world consequences, such as misinformation in chatbot responses, errors in medical image analysis and legal documents, and bias in cross-lingual information retrieval~\cite{varshney2024investigatingaddressinghallucinationsllms, christen-etal-2024-resolving, medical_negation}.

This challenge is further compounded by the fact that negations are not uniformly realized across languages. Different languages have developed diverse syntactic and morphological constructions for expressing negation~\cite{miestamo2007negation}. This variability complicates natural language processing (NLP), vision-language models (VLMs), audio (speech) foundation models, and multimodal foundation models, which already contend with an under-representation of negations in most existing benchmarks~\cite{hossain2020analysis, safavi2021negater}. As illustrated in Fig. \ref{fig:fig1}, existing text-to-image generative models often fail to interpret negations correctly, producing inaccurate outputs.\\

\noindent\textbf{Related Work:} Negation poses a significant challenge across various domains, ranging from psycholinguistics to multimodal foundation models. Psycholinguistic studies have demonstrated the cognitive complexity of negation processing, including the role of working memory, attention, and inhibitory processes. These foundational insights are informing recent advancements in NLP, VLMs, and multimodal frameworks. We now present a brief overview of the literature in this domain, focusing on psycholinguistic studies, negation understanding in language models, and its implications for multimodal and audio foundation models.

\vspace{6pt}
\noindent\textit{Psycholinguistic Studies on Processing Negations -} Research on LLM evaluation is partly inspired by psycholinguistic studies~\cite{ettinger2020bert} exploring how humans process words contextually. Some investigations have used EEG signals to understand human reactions to language stimuli~\cite{alarcao2017emotions}. Although humans acquire familiarity with negation early in life~\cite{pea1978development}, negative sentences generally demand more cognitive effort than affirmative ones~\cite{greco2020syntax, orenes2021looking}. Notably, the polarity of a statement (affirmative or negative) interacts with its truth value (true or false); while verifying true affirmatives is straightforward, verifying true negatives is found to be more difficult~\cite{khemlani2012theories}.

Psycholinguists propose \textit{composite} and \textit{interpretive} theories to explain negative comprehension~\cite{wang2021verifying}. Composite theories posit that negated concepts are a combination of parts and are processed in a two-step process — focusing first on the argument and then on its negation, whereas interpretive theories use truth-functional meanings to infer context-specific conclusions. Researchers have observed that vision-language tasks (e.g., sentence-picture verification) often mirror this two-step approach. Hence, one potential solution to improving negation understanding is to provide models with more ``time to think,'' using techniques such as chain-of-thought prompting~\cite{wei2022chain} and \texttt{PAUSE} tokens~\cite{goyal2024think} during training. However, these methods alone do not guarantee a generalizable grasp of negations; in-depth benchmarking is necessary to evaluate true reasoning capabilities.

\vspace{6pt}
\noindent \textit{Negation Understanding in LLMs:} Despite their impressive syntactic and semantic capabilities, various studies have shown that LLMs struggle to interpret negations when tested on well-crafted benchmarks for natural language inference tasks~\cite{garcia-ferrero-etal-2023-dataset}. Foundation models, extensively pretrained on unlabeled text, often fail to assign contrasting labels to affirmative and negative statements. 
Larger models seem to be more insensitive to negations, however, instruction tuning can lead to better outcomes.
They also perform almost randomly in synonym/antonym classification, emphasizing gaps in their lexical semantic knowledge of negation~\cite{truong-etal-2023-language}. A commonly cited explanation is the under-representation of negations in training corpora, since humans naturally favor affirmative expressions~\cite{ettinger2020bert}. However, simply augmenting fine-tuning data with more negative samples does not necessarily improve negation comprehension~\cite{garcia-ferrero-etal-2023-dataset}, as models often rely on superficial cues rather than genuinely internalizing negation semantics.

\vspace{6pt}
\noindent \textit{Grasping Negations in Vision-Language Contexts:} In multimodal training scenarios, such as vision-language alignment, the inclusion of visual information offers additional context. However, the fundamental limitations in language understanding persist. Contrastive learning methods, including CLIP~\cite{CLIP}, can misinterpret negations by treating visual concepts in isolation rather than capturing their relational semantics~\cite{negclip}. In one of the early works, negation-aware video retrieval was addressed by repurposing existing datasets and introducing a novel training approach. This work demonstrated a significant enhancement in models' ability to handle negated natural-language queries, leading to improved performance on standard benchmarks \cite{10.1145/3503161.3547968}. More recently, Patel et al. \cite {patel2024tripletclip} introduce TripletCLIP, a multimodal framework that enhances the understanding and representation of negation by leveraging triplet-based contrastive learning to better align textual and visual modalities. However, there is still a pressing need for more robust approaches that integrate negation-specific reasoning and representation—particularly for tasks where recognizing the absence of an object (e.g., ``not present'' or ``excluded'') is just as critical as identifying its presence.

\vspace{6pt}
\noindent \textit{Negations in Audio Foundation Models:} Despite significant advancements in audio foundation models, as evidenced by recent works~\cite{deshmukh2024pengiaudiolanguagemodel, ghosh2024gamalargeaudiolanguagemodel, DBLP:conf/acl/CaffagniCBMS0CC24, DBLP:conf/emnlp/ZhangLZZWZQ23}, negation remains largely unexplored. Effective negation interpretation is crucial across various applications, including speech recognition, translation, sentiment analysis, and conversational AI. For example, in multilingual speech recognition, accurately interpreting negations like ``not’’ in phrases such as ``I am not going’’ across diverse languages is essential for achieving correct transcriptions. In the domain of speech translation, mistranslating negations such as ``I don't like it’’ can lead to unintended consequences. Moreover, within the context of multilingual conversational AI, precise negation interpretation is indispensable for comprehending user intent and formulating appropriate responses, particularly within intricate, multi-turn dialogues. An interesting example would be a virtual assistant that must accurately interpret a user's request like ``Don't book the flight,’’ irrespective of the language spoken. However, processing negation in audio data presents a unique set of challenges, encompassing variations in prosody, intonation, acoustic ambiguity, and contextual cues across different languages. \\

\noindent \textbf{Contributions:} In this perspective, we address two crucial priorities for improving multimodal foundation models. First, we emphasize the urgent need to understand how these models interpret and manage negation across a variety of languages globally, acknowledging that many existing methods overlook important linguistic and cultural distinctions. Second, we propose a taxonomy of negation constructs that can impact multimodal foundation models. Furthermore, we advocate for the development of specialized benchmarks to rigorously measure a model’s ability to handle negation, thereby uncovering challenges and opportunities that conventional evaluations may overlook. Although models continue to make impressive strides in syntactic and semantic understanding, they frequently fail to grasp the absence or contradiction of concepts. This oversight significantly limits their reliability in applications such as text-to-image generation, audio/video retrieval, and generation. By introducing more targeted benchmarks, researchers can identify shortcomings in negation comprehension, compare efficacy in both monolingual and multilingual settings, and refine training methodologies to integrate reasoning grounded in negation. Enhancing negation understanding can also improve other desirable model properties, such as compositional and logical reasoning. Moreover, we propose research questions designed to equip multimodal foundation models with flexible, context-aware abilities to discern both the presence and absence of concepts.

\section{Taxonomy of Diversity in Negations}
\begin{figure}[h]
\centering
\includegraphics[width=\textwidth]{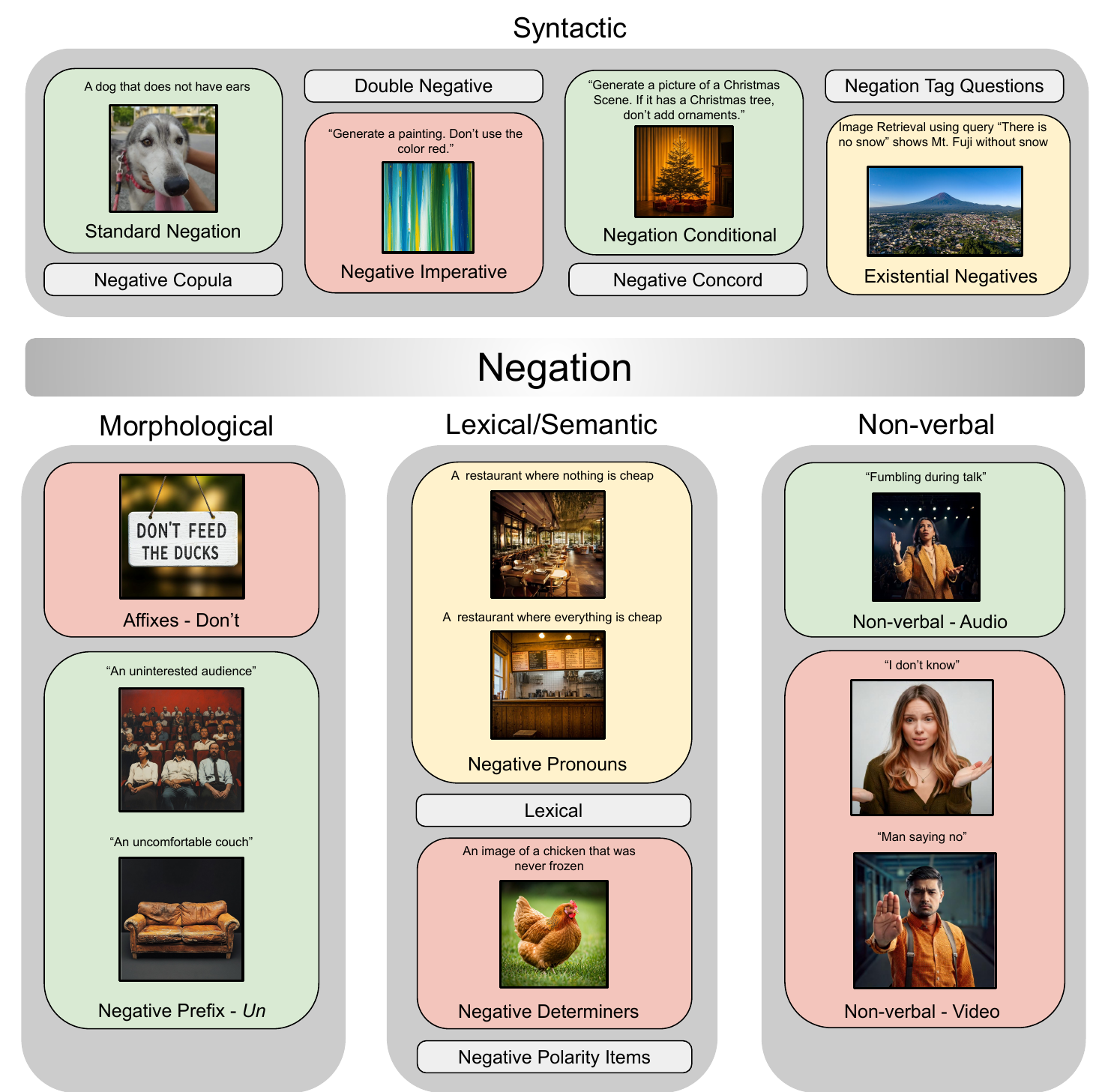}
           \caption{Overview of the proposed taxonomy for negations in multimodal contexts. The accompanying images, generated using multimodal models such as Gemini (v1.5 Flash) and Adobe Firefly, alongside examples retrieved from Google, illustrate how negated concepts are represented across different modalities in response to specific prompts, queries, or descriptions. These visuals aim to provide insight into the diverse ways negations manifest but do not capture the full complexity of the concepts. Additionally, some examples may inadvertently reflect biases present in the data or models.}
           \label{fig:fig2}
\end{figure}

Developing a clear and comprehensive taxonomy of negation across languages is crucial for identifying the specific challenges that diverse negation forms pose to multilingual multimodal foundation models, which must interpret and generate content across various modalities, including text, image, audio, and video. By systematically categorizing these negation types, researchers can devise targeted benchmarks that rigorously evaluate a model's capacity to handle subtle linguistic nuances. Moreover, a well-defined taxonomy provides a structured basis for refining both model training and evaluation methodologies, enabling practitioners to incorporate more sophisticated negation instances into datasets, metrics, and diagnostic tools. This, in turn, empowers models to navigate the complexities of negation across the diverse linguistic landscape. Building upon the cross-linguistic typology of negation~\cite{miestamo2007negation}, as shown in Fig.~\ref{fig:fig2}, we propose a taxonomy of negation for foundation models, grouping them into four overarching categories: (i) \textit{Syntactic Negations}, (ii) \textit{Morphological Negations}, (iii) \textit{Lexical and Semantic Negations}, and (iv) \textit{Prosodic, Paralinguistic, and Pragmatic Negation}. Each of these categories addresses distinct linguistic dimensions of negation and presents its own set of research opportunities in foundation models. Next, we illustrate sixteen representative negation sub-types—ranging from standard (sentential) negation to non-verbal forms. \\

\noindent\textbf{1. Syntactic Negations:} This category encompasses negations formed by adjusting a sentence’s structure to convey a negative meaning. Languages achieve this through methods such as inserting negation particles, introducing specialized words, or rearranging existing syntactic elements. Because the negative sense arises from each language’s core grammar, these approaches can vary dramatically across linguistic contexts, reflecting diverse rules and usage patterns. Within syntactic negations, several subtypes emerge—each employing distinct structural strategies to express negation.

\begin{itemize}
    \item \textit{Standard (Sentential) Negation:}  It is the most common form of negation. It typically involves a negation particle or word added to the sentence to indicate a negative meaning. In English, for instance, negation is usually marked by the word ``not,'' as in the sentence \textit{I am not going}. Many languages follow a similar pattern, though the exact words and sentence structures vary. In Romance languages like French and Spanish, negation typically involves specific particles that bracket the verb. For example, in French, the negation is often expressed by \textit{ne... pas}, as in \textit{Je ne vais pas} (I am not going). Spanish uses the word \textit{no} before the verb, such as \textit{No voy} (I am not going). These languages follow a predictable structure, but variations in how negation is expressed become more pronounced when considering non-Indo-European languages.
    \item \textit{Double Negation:} In some languages, two negation markers are used in the same sentence, a structure known as double negation. Double negation is not the same as a ``positive'' in many languages; instead, it reinforces the negative meaning. This is seen in languages like Russian, where \textit{Ya ne nikogda ne skazal} translates literally to ``I never didn’t say that,'' which simply means ``I never said that.'' Similarly, in French, \textit{Je ne fais rien} (I do nothing) involves the negation particles \textit{ne} and \textit{rien} together to reinforce the negative. Double negation exists in various languages, but its interpretation can vary. In some cases, like in African American Vernacular English (AAVE), double negation is a regular grammatical feature that intensifies the negative meaning, such as \textit{I ain’t got no money}. This structure, though grammatically different from Standard English, is a common and valid form of negation in this dialect. 
    \item \textit{Negative Concord:} It occurs when multiple negative elements appear in a sentence, but rather than canceling each other out, they collectively express a single negation. This form of negation is common in Romance and Slavic languages. In Spanish, for example, \textit{No vi a nadie} means ``I didn’t see anyone,'' where both \textit{no} and \textit{nadie} (nobody) contribute to the negative meaning. Similarly, in Italian, \textit{Non ho detto niente} (I didn’t say anything) includes the negation particle \textit{non} and the word \textit{niente} (nothing), which work together to negate the sentence. Negative concord poses challenges for models processing multilingual data because the multiple negation markers may confuse systems not specifically trained to recognize this linguistic feature. In English, multiple negatives typically create a positive, making this distinction even more crucial for systems handling multilingual tasks.
    \item \textit{Negative Copula:} In some languages, negation is expressed through special copula forms that indicate negation directly, often replacing the positive copula. Arabic, for example, uses the word \textit{laysa} to negate the verb ``to be,'' as in \textit{Ana laysa hunaka} (I am not there). This is different from many Indo-European languages that negate sentences by adding a particle, such as ``not'' or ``no,'' to the verb phrase. Japanese offers another interesting case of negative copula use. Instead of \textit{desu} (is), the negative copula form \textit{ja arimasen} is used to negate the verb, as in \textit{Kore wa hon ja arimasen} (This is not a book). This structure provides a distinct challenge for VLM models, as they need to recognize these negation-specific verb forms, which may not exist in all languages.
    \item \textit{Negation in Tag Questions:} In many languages, negation is embedded in tag questions, where the question is posed in a way that seeks confirmation or contradiction. In English, a common example is, \textit{You’re coming, aren’t you?} where the negative tag ``aren’t'' contrasts with the positive assertion. In Hindi, a similar structure exists: \textit{Tum aaoge, nahi?} (You will come, right?). Tag questions represent an interaction between affirmation and negation, requiring models to not only recognize negation but also understand the underlying pragmatics of confirmation-seeking.
    
    \item \textit{Negative Imperatives (Prohibitives):} These are used to express commands forbidding an action. In English, this takes the form of \textit{Don’t go!}, while in Korean, the phrase \textit{Hajima!} serves the same purpose, directly translating to ``Don’t do it!'' These forms of negation are often distinct from other types because they involve imperatives rather than declarative statements. Prohibitive forms add another layer of complexity for models since they often involve negation in combination with verbs of action or desire, which may not follow the same syntactic rules as standard declarative negation.
    \item \textit{Negation in Conditionals:} This kind of negation often takes on special forms, depending on the language. For example, in English, \textit{If you don’t come, I’ll be sad} contains a negation within the conditional clause. In Portuguese, a similar structure exists: \textit{Se você não vier, eu ficarei triste} (If you don’t come, I’ll be sad). Conditional negation may require models to handle both the conditional structure and the negation simultaneously, which can be particularly difficult when processing complex, multi-clause sentences.
    \item \textit{Negative Existentials:} Some languages express negation specifically through existential constructions, using dedicated words or phrases to indicate the non-existence of something. In English, \textit{There is no water} is an example of negative existentials, while in Japanese, \textit{Mizu ga arimasen} (There is no water) serves the same purpose. Negative existentials can vary greatly in their structure, requiring models to understand context-specific verbs and particles that mark non-existence, often without the use of a traditional negation particle like ``not.''
\end{itemize}

\noindent\textbf{2. Morphological Negation:} In this category, languages express negation by attaching affixes—such as prefixes or suffixes—to a word, rather than using standalone negation particles. This approach integrates negation directly into the word's structure, allowing for more compact expressions. This can also enhance communication efficiency and clarity, allowing text within VLMs to convey negation in a more integrated and seamless manner. Within morphological negation, we distinguish two main subtypes:

\begin{itemize}
   \item \textit{Affixal Negation}: Many languages express negation through morphological processes on verbs or other word classes. For example, in Turkish, the negation suffix \textit{-me} is added to verbs to create a negative meaning, as in \textit{Gitme} (Don’t go). While this method is succinct, it demands that models recognize how affixes operate across diverse linguistic systems—particularly in languages with complex morphology that may diverge from the syntax-driven structures commonly seen in English.
   
    \item \textit{Negative Prefixes:} This subtype involves the use of prefixes attached to the beginning of a word to indicate negation. In English, common negative prefixes include \textit{un-}, \textit{in-}, \textit{dis-}, and \textit{non-}, as seen in words like \textit{uncertain}, \textit{incomplete}, \textit{dislike}, and \textit{nonexistent}. German also employs negative prefixes such as \textit{un-}, as in \textit{unmöglich} (impossible). Although these constructions are typically straightforward, they can introduce ambiguity—especially when multiple prefixes share similar spellings but hold different meanings, or when context alters their significance. Recognizing and correctly interpreting these prefixes remains a key challenge in multilingual and morphologically rich environments.
\end{itemize}

\noindent\textbf{3. Lexical and Semantic Negation:} This category embeds negation within specific words or relies on the inherent semantic properties of certain lexical items. Unlike syntactic negations, which modify sentence structure, lexical and semantic negations operate at the word level. This approach allows for more nuanced and compact expressions of negation. Understanding this category is vital to capturing how negation emerges from a language’s lexicon and semantics, rather than from explicit grammatical constructions. Within lexical and semantic negations, we identify three distinct subtypes, each posing unique challenges for multilingual foundation models.

\begin{itemize}
    \item \noindent \textit{Negative Pronouns and Determiners:} Certain languages possess dedicated pronouns or determiners that inherently convey negative meanings, eliminating the need for separate negation particles. For example, in English, pronouns such as \textit{nobody} and \textit{nothing} intrinsically negate the subject or object of a sentence, as seen in \textit{Nobody came to the party} or \textit{I have nothing to say}. Similarly, German utilizes the negative determiner \textit{kein}, as in \textit{Ich habe kein Geld} (I have no money), to directly negate the noun it precedes. While these forms enable concise negation, they demand that models recognize and interpret inherently negative words in various contexts, rather than simply relying on standalone negation markers.

    \item \textit{Lexical Negation:} This subtype encompasses specific words that inherently carry negative meanings without requiring additional negation particles or morphological alterations. Examples in English include words like \textit{absent} (instead of ``not present'') and \textit{unavailable} (instead of ``not available''). Such negations are embedded within the lexical semantics of these words, meaning they do not always display overt syntactic markers. Consequently, models must develop a deeper understanding of word semantics and the ability to identify negative meanings embedded within individual lexical items, necessitating more sophisticated training approaches that go beyond surface-level negation detection.

    \item \noindent \textit{Negative Polarity Items (NPIs):} These are words or phrases permissible only in negative contexts. In English, for example, the word \textit{ever} in \textit{I haven’t ever seen that} qualifies as an NPI. Similarly, in Hindi, the word \textit{kabhi} functions as an NPI, appearing in negative sentences like \textit{Mai kabhi nahi gaya} (I never went). For models to correctly identify NPIs, they must not only detect negation but also recognize the specific lexical elements that depend on it for grammatical validity. The variation of NPIs across languages adds yet another layer of complexity for multilingual systems.
    
\end{itemize}

\noindent\textbf{4. Prosodic, Paralinguistic, and Pragmatic Negation:} Certain languages convey negation primarily through context, intonation, gestures, or dialectal usage, employing both prosodic and paralinguistic features to express denial or contradiction. These non-verbal and pragmatic forms of negation in audio and video modalities often defy standardized linguistic rules and grammatical structures, posing unique challenges for automated systems that typically rely on text-based cues. This category encompasses negation conveyed through sound patterns, such as variations in pitch, tone, rhythm, and stress, as well as non-verbal auditory cues like gestures and facial expressions. Understanding these nuanced forms of negation is crucial for developing multimodal foundation models capable of accurately interpreting negated statements in both audio and video contexts, ensuring reliable performance across diverse linguistic and cultural settings. There are three subtypes that exemplify this category’s richness:

\begin{itemize}
    \item \textit{Non-verbal Negation:} In some cultural and linguistic traditions, negation is expressed through non-verbal means, such as head shakes, specific hand gestures, facial expressions, or subtle shifts in intonation, rather than explicit verbal markers. For example, certain Native American languages employ a head shake to indicate negation, even though the spoken sentence structure itself may remain affirmative. Similarly, Dravidian languages like Tamil sometimes embed negation in intonational shifts that are difficult to capture without multimodal data (e.g., audio or video), highlighting the need for more holistic approaches to language modeling.
    
    \item \textit{Prosodic Negation:} Prosodic negation involves the use of variations in pitch, tone, rhythm, and stress to convey negation in spoken language. These prosodic features play a pivotal role in signaling the presence of negation, influencing how negated statements are perceived and understood by foundation models. For example, in Mandarin Chinese, tone and pitch variations can signal negation, as seen in the sentence ``\begin{CJK*}{UTF8}{gbsn}你不去吗?\end{CJK*}'' (Nǐ bù qù ma?), where the tonal shift on ``\begin{CJK*}{UTF8}{gbsn}不\end{CJK*}'' (bù) emphasizes the negation.
    
    \item \textit{Dialectal and Acoustic Variations:} This subtype addresses the challenges arising from variations in speech patterns, accents, and pronunciation that can obscure negation markers. These variations are particularly prevalent in multilingual contexts where pronunciation rules differ significantly across languages. For instance, in Spanish, regional accents may alter the pronunciation of negation particles like ``no,'' potentially affecting the model's ability to accurately interpret statements such as ``No quiero'' (I do not want). Additionally, in video contexts, acoustic ambiguity can be compounded by non-verbal cues such as facial expressions or gestures that either reinforce or contradict the spoken negation, requiring models to integrate multimodal information for accurate interpretation.
\end{itemize}

This taxonomy highlights that negation in language is a highly varied and complex phenomenon. From standard negation to double negation, negative concord, and morphological negation, languages offer a multitude of ways to express denial or contradiction. By acknowledging this diversity and developing strategies that address syntax, semantics, and pragmatics, researchers can refine multimodal foundation models to perform more accurately across the world’s vast array of linguistic contexts. Such efforts not only foster more reliable models but also contribute to the broader goal of creating inclusive, culturally competent technologies that genuinely understand and respect the diverse global linguistic landscape.

\begin{figure*}[]
\centering
\includegraphics[width=\linewidth]{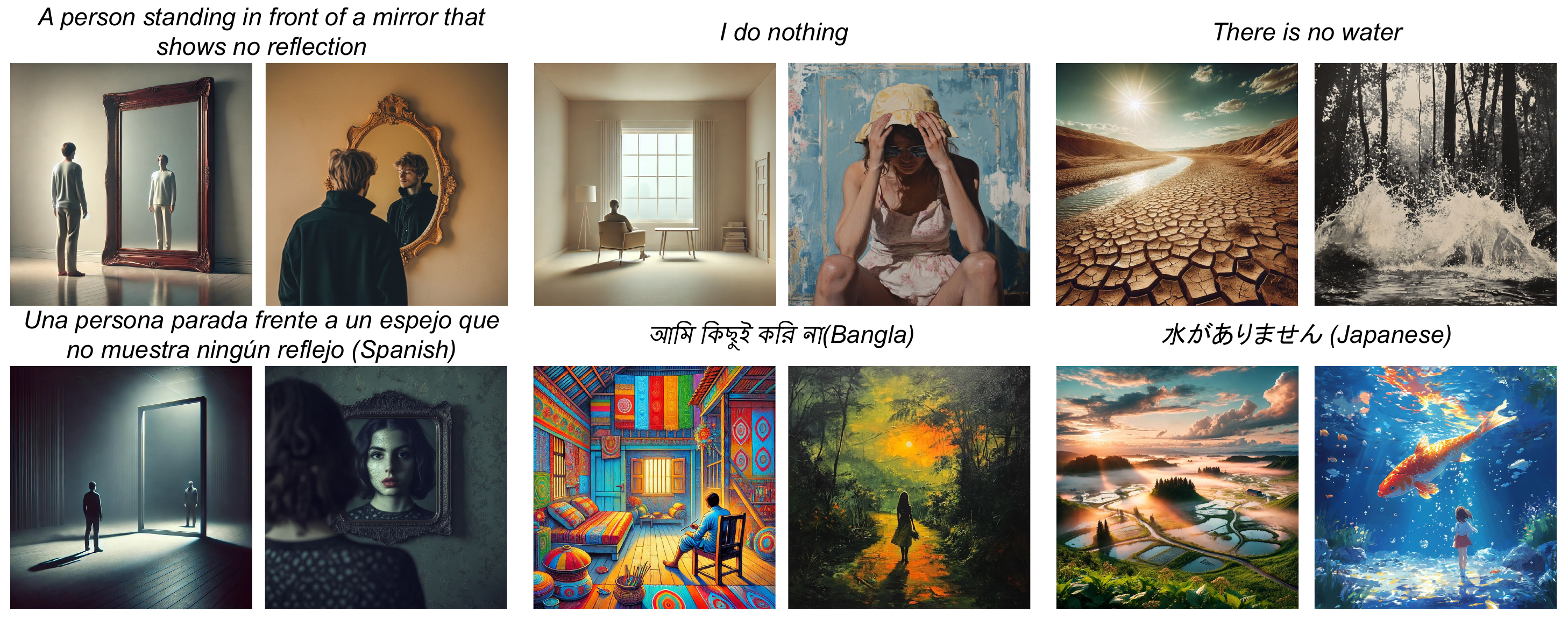}
\caption{Example showcasing how text-to-image models fail while creating images using negative prompts in different languages. The top row contains images generated through English prompts. The bottom row contains images corresponding to similar prompts in Spanish, Bangla, and Japanese. DALL-E 3 and Midjourney models are used for generation, illustrating the limitations of these models in accurately interpreting multilingual prompts and complex negations across languages.}
\label{fig:vizabstractsmall}
\end{figure*}

\section{Research Challenges and Opportunities}

As shown in Figure~\ref{fig:vizabstractsmall}, multilingual multimodal foundation models face significant challenges due to the diverse ways negations are expressed across languages, compounded by the scarcity of high-quality training data containing a wide range of negated expressions. The human tendency to focus predominantly on positive information further exacerbates this issue, leading to an underrepresentation of negated concepts in existing datasets. While standard negation in English, such as using ``not,'' may appear straightforward, languages like French and Turkish employ structures such as double negation or affixal negation, which can confuse models unfamiliar with these patterns. Negative concord, where multiple negative markers collectively express a single negation, poses particular difficulties in multilingual tasks, as foundation models may erroneously interpret these statements as positive. Lexical negation, exemplified by words like ``unhappy'' or ``absent,'' introduces additional complexity, requiring models to accurately identify and interpret inherently negative meanings embedded within the vocabulary. Non-verbal negation, conveyed through gestures like head shaking or shifts in intonation in certain cultures, further challenges multimodal models, particularly in cross-cultural contexts where such cues vary significantly. Context-specific forms of negation, such as negative polarity items and conditional clauses, demand nuanced interpretation to avoid misrepresentation. Without robust strategies to address these varied manifestations of negation, foundation models remain vulnerable to errors in critical applications, including image-text matching, translation, and sentiment analysis.

Addressing these challenges is essential for enhancing the adaptability and resilience of foundation models. Below, we propose key research questions that warrant investigation to advance model performance in multilingual and multimodal settings. Although these considerations do not exhaust the scope of future research, they illuminate the pressing issues that must be resolved to build more comprehensive, culturally sensitive foundation models.

\begin{itemize}
  
  \item \textbf{RQ1:} \emph{How do language-specific structures and semantics of negation affect foundation model performance, and to what extent does linguistic diversity influence robustness in negation understanding across various modalities?} This question seeks to uncover performance gaps in tasks such as captioning, retrieval, audio transcription, and video analysis, emphasizing the impact of cross-lingual and cross-modal differences.

   \item \textbf{RQ2:} \emph{What misinterpretations commonly arise in foundation models when handling language-specific negation patterns across different modalities?} This question seeks to identify systematic challenges, such as the misinterpretation of doubly negated sentences as affirmative or misunderstanding negation in video and audio contexts, which can undermine accurate negation handling across diverse linguistic and modal settings.
    
    \item \textbf{RQ3:} \emph{In what ways do cultural contexts shape foundation models’ interpretation of negation across multilingual and multimodal scenarios, and what adaptations are necessary for context-aware performance?} This question investigates how cultural nuances and localized usage patterns influence negation handling in text, images, videos, and audio, highlighting the modifications needed for truly context-aware models.

    \item \textbf{RQ4:} \emph{How do double or nested negations (e.g., ``not uncommon'') across languages impact foundation model performance across different modalities, and what techniques could enhance their interpretation?} This question explores the complexities posed by layered negation constructs in text, image captions, and audio descriptions, examining strategies to improve model accuracy in interpreting such expressions.
    
    \item \textbf{RQ5:} \emph{How do negations within idiomatic or figurative expressions influence foundation model performance across different modalities, and which training strategies can effectively manage these subtleties?} This question examines the challenges arising from non-literal negations in text, image descriptions, and video/audio content, assessing how models can better capture figurative language and idiomatic turns of phrase.    
    \item \textbf{RQ6:} \emph{Is a universal negation-handling framework viable for foundation models, or must strategies be tailored to individual languages and modalities for optimal outcomes?} This question weighs the trade-offs between developing a one-size-fits-all approach and designing language- and modality-specific solutions, aiming to clarify the most effective strategies for robust negation handling.  
    
    \item \textbf{RQ7:} \emph{Which training methodologies best accommodate language-specific negation structures across different modalities and improve overall foundation model resilience?}  
    This question proposes improvements to training protocols and data augmentation methods that can more effectively account for diverse negation patterns in different modalities.
    
    \item \textbf{RQ8:} \emph{How do multimodal inputs (e.g., text, images, video, audio) affect negation interpretation in foundation models, and how can models leverage these modalities for better performance?} This question evaluates the role of cross-modal cues in enhancing negation handling, focusing on the synergy between textual, visual, and auditory signals.

    \item \textbf{RQ9:} \emph{Can transfer learning be leveraged to generalize negation handling in under-resourced languages and modalities with limited training data?} This question explores how transfer learning might facilitate better negation processing in low-resource linguistic and modal settings, emphasizing scalability and adaptability.

     \item \textbf{RQ10:} \emph{How do contextual embeddings (e.g., BERT-like representations) influence negation comprehension in foundation models across different modalities, particularly for multilingual scenarios?} This question examines the interplay between contextual embeddings and language-specific negation features, investigating ways to refine embedding architectures for robust multilingual and multimodal performance.

     \item \textbf{RQ11:} \emph{In what ways do compression and noise in multimodal data (e.g., text, images, video, audio) affect negation accuracy in multilingual foundation models, and how can these factors be mitigated?} This question assesses how data quality issues in various modalities undermine negation interpretation, proposing methods to reduce errors introduced by noisy or compressed inputs.    
\end{itemize}

\section{Discussion and Future Directions}
Initial efforts in advancing negation understanding have been encouraging, exemplified by datasets like \textit{NeQA}~\cite{zhang-etal-2023-beyond}, \textit{This-is-not-a-Dataset}~\cite{garcia-ferrero-etal-2023-dataset}, \textit{SugarCrepe}~\cite{hsieh2023sugarcrepe}, \textit{CC-Neg}~\cite{singh2024learn}, and studies on bias in Indic languages \cite{surbhieccv2024}. Moreover, research has shown promise in various domains, including the medical and legal fields~\cite{christen-etal-2024-resolving, medical_negation}. However, these advances only begin to address the multifaceted nature of negation across diverse linguistic and multimodal contexts. We posit that future research should integrate specialized negation benchmarks and architectural-level enhancements, such as language-specific strategies, specialized attention mechanisms, and multimodal hybrid architectures, to more comprehensively capture and interpret negation.

\vspace{6pt}

\noindent\textbf{Specialized Negation Benchmarks:} Establishing dedicated benchmarks for negation-specific constructs across monolingual, multilingual, unimodal, and multimodal foundation model settings is crucial to enhancing negation handling in large foundation models. These benchmarks should capture the full linguistic diversity of negation forms, ranging from double negation and negative concord to affixal negation and non-verbal cues, thereby assessing how models interpret and integrate negation across different languages and modalities. To facilitate fair comparisons, key evaluation metrics must be standardized while encompassing a wide range of language families, dialects, and cultural contexts. Furthermore, incorporating tasks that mirror real-world applications—such as text-to-image retrieval, audio-based question answering, unimodal text generation, and multilingual content understanding—enables a more practical and comprehensive assessment of model performance in complex and varied scenarios. By developing such specialized benchmarks, researchers can systematically evaluate and improve the ability of foundation models to comprehend and handle negation, ensuring their reliability and effectiveness across diverse linguistic and multimodal environments.

\vspace{6pt}
\noindent\textbf{Integrative Architectural Approaches for Negation:} To enhance the understanding of negation in multimodal foundation models, several architectural advancements are essential. \textit{Language-specific tokenization and preprocessing}~\cite{limisiewicz-etal-2023-tokenization, chelombitko2024qtok} play a pivotal role by ensuring that critical negation markers are accurately captured across diverse linguistic variations. By introducing targeted tokenization schemes tailored to specific languages, models can better recognize negation structures such as negative concord, double negation, and unique morphological markers, especially when supported by a comprehensive multilingual corpus. Complementing this, the incorporation of \textit{fine-grained attention mechanisms}~\cite{choi2018fine} dedicated to detecting and resolving the scope of negation can significantly boost model performance. These specialized attention networks focus on specific sentence segments where negation occurs, effectively managing nested negations and multiple markers that may reinforce or negate meanings, thereby reducing the likelihood of misclassifying nuanced instances of denial or contradiction. Furthermore, advancing the model architecture through a multimodal hybrid framework that includes a fusion layer capable of processing textual, visual, and auditory inputs can dramatically improve negation comprehension. Many languages and cultures utilize non-textual indicators such as gestures and intonation to convey negation; by integrating these additional data channels, multimodal foundation models can interpret negation more holistically, regardless of whether it appears in written form, spoken utterances, or visual cues. 

\bibliography{aaai25}

\end{document}